\documentclass[10pt,twocolumn,letterpaper]{article}

\usepackage{cvpr}              

\definecolor{cvprblue}{rgb}{0.21,0.49,0.74}
\usepackage[pagebackref,breaklinks,colorlinks,citecolor=cvprblue]{hyperref}

\def\paperID{*****} 
\def\confName{CVPR}
\def\confYear{2024}

\usepackage{blindtext}
\usepackage{listings}
\usepackage{pgfplotstable}
\usepackage{enumitem}

\newcommand{\mcrot}[4]{\multicolumn{#1}{#2}{\rlap{\rotatebox{#3}{#4}~}}} 

\begin{document}

\title{CausalPlayground: Addressing Data-Generation Requirements in Cutting-Edge Causality Research}

\author{Andreas Sauter\\
Vrije Universiteit Amsterdam\\
De Boelelaan 1111, 1081 HV Amsterdam\\
{\tt\small a.sauter@vu.nl}
\and
Erman Acar\\
Universiteit van Amsterdam\\
Science Park 900, 1098 XH Amsterdam\\
{\tt\small e.acar@uva.nl}
\and
Aske Plaat\\
Leiden University\\
Einsteinweg 55, 2333 CC Leiden\\
{\tt\small 
a.plaat@liacs.leidenuniv.nl}
}

\maketitle

\begin{abstract}
Research on causal effects often relies on synthetic data due to the scarcity of real-world datasets with ground-truth effects. Since current data-generating tools do not always meet all requirements for state-of-the-art research, ad-hoc methods are often employed. This leads to heterogeneity among datasets and delays research progress. We address the shortcomings of current data-generating libraries by introducing \emph{CausalPlayground}, a Python library that provides a standardized platform for generating, sampling, and sharing structural causal models (SCMs). \emph{CausalPlayground} offers fine-grained control over SCMs, interventions, and the generation of datasets of SCMs for learning and quantitative research. Furthermore, by integrating with  \emph{Gymnasium}, the standard framework for reinforcement learning (RL) environments, we enable online interaction with the SCMs. 
Overall, by introducing \emph{CausalPlayground} we aim to foster more efficient and comparable research in the field. All code and API documentation is available at \url{https://github.com/sa-and/CausalPlayground}.
\end{abstract}

\section{Introduction}\label{sec:intro}
\begin{table*}[t]
\centering
\begin{tabular}{lccccccc}
& \mcrot{1}{l}{30}{interventional data (R1)} & \mcrot{1}{l}{30}{interaction (R2)} & \mcrot{1}{l}{30}{SCM-controlled (R3)} & \mcrot{1}{l}{30}{SCM-Generation (R4)} & \mcrot{1}{l}{30}{DAG-Generation} & \mcrot{1}{l}{30}{API-Documentation} & \mcrot{1}{l}{30}{Language} \\\hline
Do-Why {\cite{Sharma2020DoWhy:Inference}} & \checkmark & X & X & X & X & \checkmark & P \\
TETRAD {\cite{Scheines1998TheSpecification,Ramsey2023Py-TetradSearch}} & X & X & X & X & \checkmark & \checkmark & J/P/R\\
cause2e {\cite{DanielGrunbaum2021Cause2e:Analysis.}} & \checkmark & X & X & X & \checkmark & \checkmark & P \\
Lawrence et al. \cite{Lawrence2021DataData} & X & X & $\sim$ & X & \checkmark & X & P\\
MANM-CS {\cite{Huegle2021MANM-CS:Data}} & \checkmark & X & $\sim$ & X & \checkmark & X & P\\
BNlearn { \cite{Taskesen2020LearningPackage.}} & $\sim$ & X & X & X & X & \checkmark & P\\
causaldag {\cite{ChandlerSquires2018Causaldag:Models}} & $\sim$ & X & X & X & \checkmark & \checkmark & P\\
gCastle {\cite{Zhang2021GCastle:Discovery}} & X & X & $\sim$ & X & \checkmark & \checkmark & P \\
CDT {\cite{Kalainathan2019CausalPython}} & X & X & $\sim$ & X & \checkmark & \checkmark & P \\
SCModels {\cite{Aichmuller2023StructuralModels}} & \checkmark & X & \checkmark & X & X & X & P\\
R6causal {\cite{Karvanen2023R6causal:Models}} & \checkmark & X & \checkmark & X & X & \checkmark & R \\
CausalWorld {\cite{Ahmed2020CausalWorld:Learning}} & \checkmark & \checkmark & X & X & X & \checkmark & P\\
SynTReN {\cite{VandenBulcke2006SynTReN:Algorithms}} & X & X & X & X & \checkmark & X & J\\
pcalg {\cite{AlainHauser2012CharacterizationGraphs}} & X & X & X & X & \checkmark & \checkmark & R\\
JustCause {\cite{Franz2020JustCause}} & X & X & \checkmark & X & X & \checkmark & P \\
\textbf{CausalPlayground} & \checkmark & \checkmark & \checkmark & \checkmark & \checkmark & \checkmark & P\\\hline
\end{tabular}
\caption{Overview of software libraries that can be used for data-generation for causality research. Symbols \checkmark, X, and $\sim$ indicate whether a requirement is met, not met, or partially met, respectively. J, P, R refer to Java, Python, and R, respectively. We evaluate the methods on the ability to sample interventional data, interact with the causal model, control models on a functional level, generate models,  generate directed acyclic graphs (DAG), the availability of a detailed API documentation, and the programming language that it is written in.}
\label{tab:comparison}
\end{table*}

Ever since the formalization of causality \cite{Pearl2009Causality} the field has gained significant attention for improving inference from data, scientific discovery, and others. Progress in causality research relies heavily on data. However, a major challenge is the lack of real-world data with known ground truth causal relations \cite{Cheng2022EvaluationAlgorithms}. For example, even after decades of research, only a few strong real-world benchmark datasets are available for causal structure discovery \cite{Cheng2022EvaluationAlgorithms,Vowels23yall}. Consequently, many state-of-the-art methods use synthetically generated data, such as in causal representation learning \cite{Lippe2023BISCUIT:Interactions}, causal discovery \cite{Deleu2023JointNetwork}, among others. 
 
Naturally, every research question comes with its specific requirements for the data-generating process that is being investigated. In causality particularly though, some requirements can be observed that are common among many instances. We identify them as requirements for a causal data-generating library to be useful for a broad variety of research questions as follows:

\begin{itemize}
\item[\textbf{R1}] \textbf{Interventional data generation:} Intuitively, interventions are experiments in an environment that are crucial for identifying causal effects \cite{Bareinboim2022OnInference}. Therefore, a general causal data-generation framework must facilitate interventions.
\item[\textbf{R2}] \textbf{Interaction with the causal model:} It is becoming increasingly clear, that interacting with the causal model is beneficial for many tasks. For example, intervening after every sample can improve sample efficiency for causal discovery methods \cite{Sauter2023ADiscovery, Sauter2024CORE:Learning} and causal inference \cite{Toth2022ActiveInference}. This requirement is further highlighted by the popularity of interactive frameworks like Causal World \cite{Ahmed2020CausalWorld:Learning} and various interactive causal models created ad-hoc for specific research questions \cite{Zhu2019CausalLearning, Lippe2022Citris:Sequences, Toth2022ActiveInference}.
\item[\textbf{R3}] \textbf{Fine-grained control over the causal model:} As causal inference and discovery often investigate settings with clearly defined assumptions on the data, like the popular linear functions with additive non-Gaussian noise \cite{Peters2017ElementsAlgorithms} assumption, a general data-generation library must provide detailed control over the functional relations of the causal model.
\item[\textbf{R4}] \textbf{Causal model generation for quantitative results:} Recent research in causal methods, such as \cite{Lorch2022AmortizedLearning, Sauter2023ADiscovery,Deleu2023JointNetwork,Sauter2024CORE:Learning},  emphasize the need for training and evaluation not merely on a single causal model, but rather on data-sets of models. This results in the requirement for a general causal data-generating library to be able to easily create many causal models.
\end{itemize}

 As our comparison in \cref{sec:related_work} shows existing tools fall short of addressing all current requirements within one framework. This leads to many researchers having to rely on ad-hoc data generation processes, introducing undesirable heterogeneity of datasets amongst methods, more difficult comparisons of results, and generally obstructing the research progress of the field. 

 To offer a solution to this problem we created \emph{CausalPlayground}, a Python library for generating synthetic data and providing generation processes for causal models. Our library enables fine-grained control over the models, integrates processes within the popular RL framework \emph{Gymnasium} \cite{Towers2023Gymnasium} to accommodate the requirement for interacting with the causal model, and implements the generation of many causal models at once to enable learning and quantitative research. It thereby constitutes a framework for causal data generation that allows for more standardized and easy-to-share causal data-generation processes. 
 
 In the remainder of the paper we will compare current causal data-generating libraries in \cref{sec:related_work}, introduce fundamental notions of causality more formally in \cref{sec:fundamental}, provide an overview of \emph{CausalPlayground} in \cref{sec:overview}, outline a simple use-case in \cref{sec:use_case}, and give an outlook on future versions of this library in \cref{sec:conclusion}.

 \section{Related Work}\label{sec:related_work}
Based on the requirements we identified in \cref{sec:intro}, we investigate the current landscape of general purpose causal data-generation tools. We restrict our comparison to packages specifically designed for causality-related research, acknowledging that other general-purpose tools have been used, such as common RL environments like MuJoCo \cite{Todorov2012MuJoCo:Control}. 
The comparison is summarized in \cref{tab:comparison}.

 Regarding interventional data-generation (R1), many available packages, including Do-Why \cite{Sharma2020DoWhy:Inference}, cause2e \cite{DanielGrunbaum2021Cause2e:Analysis.}, MANM-CS \cite{Huegle2021MANM-CS:Data}, SCModels \cite{Aichmuller2023StructuralModels}, R6causal \cite{Karvanen2023R6causal:Models}, and CausalWorld \cite{Ahmed2020CausalWorld:Learning}, allow for sampling from interventional distributions for specifiable functional interventions. BNlearn \cite{Taskesen2020LearningPackage.} and causaldag \cite{ChandlerSquires2018Causaldag:Models} offer less versatile methods for intervening, and facilitate interventions by changing the node conditional distributions or the graphical structure, respectively.

Considering interaction with causal models (R2) to the best of our knowledge, only CausalWorld \cite{Ahmed2020CausalWorld:Learning} provides out-of-the-box functionality for actively interacting with the model, highlighting an opportunity for useful new tools to meet this requirement.

With respect to generating causal models with specific assumptions on the causal functions (R3) SCModels \cite{Aichmuller2023StructuralModels}, R6causal \cite{Karvanen2023R6causal:Models}, and JustCause \cite{Franz2020JustCause} give full control over functional relations and noise distributions, while Lawrence et al. \cite{Lawrence2021DataData}, MANM-CS \cite{Huegle2021MANM-CS:Data}, gCastle \cite{Zhang2021GCastle:Discovery}, and CDT \cite{Kalainathan2019CausalPython} provide coarse-grained control that simultaneously applies to all functions and distributions in the model.

Lastly, no general-purpose framework for causal data-generation provides methods to generate sets of causal models with specific functional relations and noise distributions (R4) out-of-the-box.

Overall, this comparison shows a gap in libraries that simultaneously address some of the most important requirements for causal data-generation. This opens the opportunity to develop a tool that is relevant to a broader set of research questions.

\section{Fundamental Concepts of Causality}\label{sec:fundamental}
In this section, we provide a condensed overview of the necessary formal concepts for our approach. Let an SCM $M$ be a tuple $M=(\mathcal{X}, \mathcal{U}, \mathcal{F}, \mathcal{P})$, where $\mathcal{X}$ and $\mathcal{U}$ are sets of endogenous and exogenous random variables, respectively, $\mathcal{F}$ is a set of functions $f_i$ that map the direct causes of $X_i \in \mathcal{X}$ to $X_i$, and $\mathcal{P}$ is a set of pairwise independent probability distributions $P_i$ s.t. $U_i$ follows distribution $P_i$ with $U_i \in \mathcal{U}$. Furthermore, we denote interventions as $do(X_i = g(...))$, where $g$ is an arbitrary function that takes as input a subset of $\mathcal{X}\cup \mathcal{U}$. Such an intervention replaces function $f_i$ with function $g$. Applying a set of $N$ arbitrary interventions $a_t = \{do_0(\ldots), \ldots, do_N(\ldots)\}$ to an SCM, means replacing all corresponding functions simultaneously. Furthermore, each SCM $M$ induces a distribution $P^M(\mathcal{X}, \mathcal{U}\mid a_t)$ that we can sample via ancestral sampling, hence the SCM can be seen as a data-generating process.

\section{CausalPlayground}\label{sec:overview}
In this section we provide an overview of the \emph{CausalPlayground} library.  By implementing the functionalities that address the requirements R1 - R4, we provide an off-the-shelf causal data-generation platform that can be used for state-of-the-art causality research. 

\subsection{Fine-grained Control Over the SCM} For each endogenous variable $X_i$, the structural equation $f_i$ can be arbitrarily defined. Similarly, the distribution of the exogenous variables $P_j$ can be arbitrarily defined. Notably, this entails the possibility of defining pixel-based environments. An example for creating the SCM with $\mathcal{F}=\{A\leftarrow U+5, Effect \leftarrow 2A$\} and $U\sim Uniform(3,8)$ is provided below:
\begin{lstlisting}[language=Python]
scm = StructuralCausalModel()
scm.add_endogenous_var('A', 
    lambda noise: noise+5, 
    {'noise': 'U'})
scm.add_exogenous_var('U', 
    random.randint, 
    {'a': 3, 'b': 8})
scm.add_endogenous_var('Effect', 
    lambda x: x*2, 
    {'x': 'A'})
\end{lstlisting}

This level of control over the SCM is a significant improvement compared to other libraries, which often provide more coarse-grained control.

\subsection{Arbitrary Interventions} 
Interventions can be applied to existing SCMs using arbitrary functions $g$.  The only restriction is that the new function does not induce cyclic causal relations. In the example code snippet:
\begin{lstlisting}[language=Python]
scm.do_interventions([("Effect", 
                       (lambda a: a+1, 
                       {'a':'A'}))])
\end{lstlisting}
the intervention sets the value of the variable \texttt{Effect} to the value of $\textit{A}+1$, using a lambda function that takes \texttt{a} as an argument and a dictionary that maps \texttt{a} to the variable name \texttt{A}. This, effectively, applies $do(Effect=A+1)$ to the SCM.

\begin{figure*}[t]
    \centering
    \includegraphics[width=0.7\textwidth]{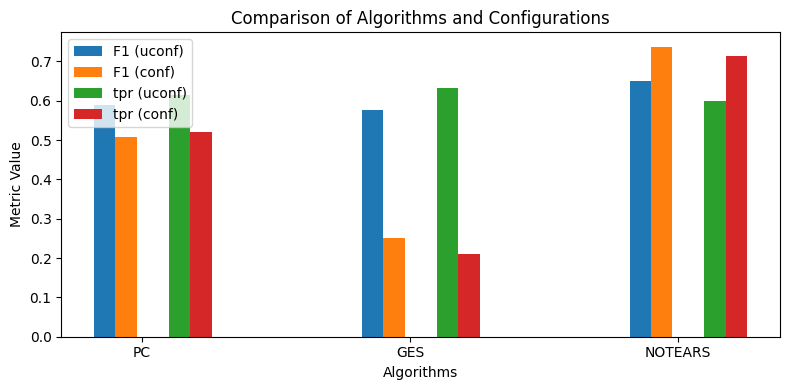}
    \caption{Experimental results of the simple causal discovery algorithm comparison use-case. F1 and tpr correspond to the F1 score and the true positive rate respectively.}
    \label{fig:results}
\end{figure*}

\subsection{Interaction with the SCM} 
Users can intervene actively in existing SCMs. At each step, a set of interventions $a_t$ can be applied, after which a sample of the endogenous and exogenous variables is drawn from the induced distribution. The interventions are undone before the next step. The interaction is enabled by wrapping the SCMs in the popular \emph{Gymnasium} environment, facilitating standardized integration with (deep) RL methods \cite{Plaat2022DeepLearning}. In the example code snippet:
\begin{lstlisting}[language=Python]
env = SCMEnvironment(scm, 
       possible_interventions=
        [("A", (lambda: 5, {})), 
         ("Effect", (lambda a: a+1, 
                     {'a':'A'}))])
\end{lstlisting}
An \texttt{SCMEnvironment} object is created by passing the \texttt{scm} object and a list of two possible interventions with the same syntax as described before. The resulting \texttt{env} object can be used to interact with the SCM using the standard Gymnasium environment interface, allowing for interactive exploration and experimentation with the causal model.

More specifically, calling \texttt{env.step(action)} applies the interventions defined in the action via their index, samples the intervened SCM, determines the new observation, termination flag, truncated flag, and reward, and, finally undoes the interventions. Importantly, the observation, termination, truncated, and reward functions can be specified by the user.

\subsection{SCM generation} 
CausalPlayground's SCM generation procedure provides a flexible approach to creating datasets of SCMs. Users can specify the number of endogenous variables and exogenous variables, the presence of confounders, and define the class of possible causal relations. This is exemplified by the following code:
\begin{lstlisting}
gen = SCMGenerator(all_functions=
                   {'linear': f_linear})
scm_unconfounded = gen.create_random(N=5,
            possible_functions=['linear'],
            n_endo=5, n_exo=4,
            exo_distribution=random.gauss,
            exo_distribution_kwargs=
              {mu=0, sigma=1},
            allow_exo_confounders=
              False)[0]
\end{lstlisting}
that creates 5 random SCMs with the causal relations following a predefined \texttt{f\_linear} function, 5 endogenous variables, 4 exogenous variables, a Gaussian exogenous distribution with $\mu = 0, \sigma =1$ for each exogenous variable, and no confounders. Similarly, an SCM can also be generated from a given causal structure.

This comprehensive control over the SCM generation process facilitates the systematic evaluation and comparison of causal models. 


\subsection{Further Implementation Details}
\emph{CausalPlayground} is implemented in Python with five main classes. The \texttt{StructuralCausalModel} class represents an SCM and provides methods for manipulation and sampling. The \texttt{SCMGenerator} creates random SCMs based on specified parameters, while the \texttt{CausalGraphGenerator} and \texttt{CausalGraphSetGenerator} generate random causal graphs and sets of unique causal graphs, respectively. The \texttt{SCMEnvironment} class, allows interaction with an SCM. Furthermore, \emph{CausalPlayground} relies on external classes such as \texttt{networkx.DiGraph}, \texttt{pandas.DataFrame}, \texttt{Gymnasium.Box}, and \texttt{Gymnasium.Sequence} for various functionalities. All code is published under the open-source MIT license and we highly welcome contributions.

\section{Use-Case: Comparison of Causal Discovery Algorithms}\label{sec:use_case}
As one of many possible use-cases, we outline how to use the \emph{CausalPlayground} library to compare different causal discovery algorithms. We generate synthetic data using the library's functions and evaluate the performance of the algorithms on both confounded and unconfounded, automatically generated SCMs. We use the causal discovery algorithms provided by the gCastle \cite{Zhang2021GCastle:Discovery} library. The detailed implementation can be found in the repository examples.

First, we generate distinct confounded and unconfounded causal graphs using the \texttt{CausalGraphSetGenerator} class from \emph{CausalPlayground}. Each of the sets has 4 endogenous variables and 4 exogenous variables. Based on these graphs we generate the SCMs for evaluation using the \texttt{SCMGenerator} class. We define the functional relations as linear additive functions and functions that add all causes and multiply the value of two randomly selected causes of a variable. More specifically, for every endogenous variable, the function is drawn from either of these two classes.

We then iterate over the 30 generated SCMs and draw 100 samples per SCM. We evaluate the performance of each causal discovery algorithm on both confounded and unconfounded datasets. More specifically, we compare PC \cite{Spirtes2000CausationSearch}, GES \cite{Chickering2003OptimalSearch}, and NOTEARS \cite{Zheng2018DAGsLearning} and measure F1 score and the true-positive rate with gCastle \cite{Zhang2021GCastle:Discovery}. The results can be seen in \cref{fig:results}.

While not providing specifically insightful results to the community, this simple use-case showcases the usefulness of \emph{CausalPlayground} for causality research due to its ability to easily generate, sample, and eventually share SCMs. 

\section{Conclusion and Outlook}\label{sec:conclusion}
In this work, we identified the core requirements for data-generation in causal research and found that they are unsatisfactorily met by the current landscape of data-generation tools. To offer a solution, we introduced \emph{CausalPlayground}, a Python library that allows for sampling of, generation of, and online interaction with SCMs. By providing functionality to address some of the main requirements of causality research, this library can enable a broad research community to easily share their SCMs and more rigorously compare its methods. 

For future versions of this library, we are anticipating the integration of more diverse causal models such as Bayesian networks, and optimizations regarding parallelization for fast deployment on GPUs. Ultimately, we envision that our proposed library contributes to faster scientific progress in fields that rely on causality. 

\section*{Acknowledgements}
We thank Floris den Hengst for lending a helpful hand when trying to fix a bug.

This research was partially funded by the Hybrid Intelligence Center, a 10-year programme funded by the Dutch Ministry of Education, Culture and Science through the Netherlands Organisation for Scientific Research, https://hybrid-intelligence-centre.nl, grant number 024.004.022

{
    \small
    \bibliographystyle{ieeenat_fullname}
    \bibliography{references}

\begin{thebibliography}{35}
\providecommand{\natexlab}[1]{#1}
\providecommand{\url}[1]{\texttt{#1}}
\expandafter\ifx\csname urlstyle\endcsname\relax
  \providecommand{\doi}[1]{doi: #1}\else
  \providecommand{\doi}{doi: \begingroup \urlstyle{rm}\Url}\fi

\bibitem[Ahmed et~al.(2020)Ahmed, Tr{\"{a}}uble, Goyal, Neitz, W{\"{u}}thrich, Bengio, Sch{\"{o}}lkopf, and Bauer]{Ahmed2020CausalWorld:Learning}
Ossama Ahmed, Frederik Tr{\"{a}}uble, Anirudh Goyal, Alexander Neitz, Manuel W{\"{u}}thrich, Yoshua Bengio, Bernhard Sch{\"{o}}lkopf, and Stefan Bauer.
\newblock {CausalWorld: A Robotic Manipulation Benchmark for Causal Structure and Transfer Learning}, 2020.

\bibitem[Aichm{\"{u}}ller(2023)]{Aichmuller2023StructuralModels}
Michael Aichm{\"{u}}ller.
\newblock {Structural Causal Models}, 2023.

\bibitem[{Alain Hauser} and {Peter B{\"{u}}hlmann}(2012)]{AlainHauser2012CharacterizationGraphs}
{Alain Hauser} and {Peter B{\"{u}}hlmann}.
\newblock {Characterization and greedy learning of interventional Markov equivalence classes of directed acyclic graphs}.
\newblock \emph{Journal of Machine Learning Research}, 13:\penalty0 2409--2464, 2012.

\bibitem[Bareinboim et~al.(2022)Bareinboim, Correa, Ibeling, and Icard]{Bareinboim2022OnInference}
Elias Bareinboim, Juan~D. Correa, Duligur Ibeling, and Thomas Icard.
\newblock {On Pearl’s Hierarchy and the Foundations of Causal Inference}.
\newblock In \emph{Probabilistic and Causal Inference}, pages 507--556. ACM, New York, NY, USA, 2022.

\bibitem[{Chandler Squires}(2018)]{ChandlerSquires2018Causaldag:Models}
{Chandler Squires}.
\newblock {causaldag: creation, manipulation, and learning of causal models}, 2018.

\bibitem[Cheng et~al.(2022)Cheng, Guo, Moraffah, Sheth, Sel{\c{c}}uk~Candan, Liu, Kong, and Liu~are]{Cheng2022EvaluationAlgorithms}
Lu Cheng, Ruocheng Guo, Raha Moraffah, Paras Sheth, K Sel{\c{c}}uk~Candan, Huan Liu, Hong Kong, and Huan Liu~are.
\newblock {Evaluation Methods and Measures for Causal Learning Algorithms}.
\newblock \emph{IEEE Transactions on Artificial Intelligence}, 3\penalty0 (6), 2022.

\bibitem[Chickering(2003)]{Chickering2003OptimalSearch}
David~Maxwell Chickering.
\newblock {Optimal Structure Identification with Greedy Search}.
\newblock \emph{J. Mach. Learn. Res.}, 3\penalty0 (null):\penalty0 507--554, 2003.

\bibitem[{Daniel Gr{\"{u}}nbaum}(2021)]{DanielGrunbaum2021Cause2e:Analysis.}
{Daniel Gr{\"{u}}nbaum}.
\newblock {cause2e: A Python package for end-to-end causal analysis.}, 2021.

\bibitem[Deleu et~al.(2023)Deleu, Nishikawa-Toomey, Subramanian, Malkin, Charlin, and Bengio]{Deleu2023JointNetwork}
Tristan Deleu, Mizu Nishikawa-Toomey, Jithendaraa Subramanian, Nikolay Malkin, Laurent Charlin, and Yoshua Bengio.
\newblock {Joint Bayesian Inference of Graphical Structure and Parameters with a Single Generative Flow Network}.
\newblock In \emph{Advances in Neural Information Processing Systems}, pages 31204--31231. Curran Associates, Inc., 2023.

\bibitem[Franz et~al.(2020)Franz, Wilhelm, and Kulkarni]{Franz2020JustCause}
Maximilian Franz, Florian Wilhelm, and Tanmay Kulkarni.
\newblock {JustCause}, 2020.

\bibitem[Huegle et~al.(2021)Huegle, Hagedorn, B{\"{o}}hme, P{\"{o}}rschke, Umland, and Schlosser]{Huegle2021MANM-CS:Data}
Johannes Huegle, Christopher Hagedorn, Lukas B{\"{o}}hme, Mats P{\"{o}}rschke, Jonas Umland, and Rainer Schlosser.
\newblock {MANM-CS: Data generation for benchmarking causal structure learning from mixed discrete-continuous and nonlinear data}.
\newblock In \emph{WHY-21 at NeurIPS 2021}, 2021.

\bibitem[Kalainathan and Goudet(2019)]{Kalainathan2019CausalPython}
Diviyan Kalainathan and Olivier Goudet.
\newblock {Causal Discovery Toolbox: Uncover causal relationships in Python}.
\newblock 2019.

\bibitem[Karvanen(2023)]{Karvanen2023R6causal:Models}
Juha Karvanen.
\newblock {R6causal: R6 Class for Structural Causal Models}, 2023.

\bibitem[Lawrence et~al.(2021)Lawrence, Kaiser, Sampaio, and Sipos]{Lawrence2021DataData}
Andrew~R. Lawrence, Marcus Kaiser, Rui Sampaio, and Maksim Sipos.
\newblock {Data Generating Process to Evaluate Causal Discovery Techniques for Time Series Data}.
\newblock 2021.

\bibitem[Lippe et~al.(2022)Lippe, Magliacane, L{\"{o}}we, Asano, Cohen, and Gavves]{Lippe2022Citris:Sequences}
Phillip Lippe, Sara Magliacane, Sindy L{\"{o}}we, Yuki~M Asano, Taco Cohen, and Stratis Gavves.
\newblock {Citris: Causal identifiability from temporal intervened sequences}.
\newblock In \emph{International Conference on Machine Learning}, pages 13557--13603, 2022.

\bibitem[Lippe et~al.(2023)Lippe, Magliacane, L{\"{o}}we, Asano, Cohen, and Gavves]{Lippe2023BISCUIT:Interactions}
Phillip Lippe, Sara Magliacane, Sindy L{\"{o}}we, Yuki~M Asano, Taco Cohen, and Efstratios Gavves.
\newblock {BISCUIT: Causal Representation Learning from Binary Interactions}.
\newblock In \emph{The 39th Conference on Uncertainty in Artificial Intelligence}, 2023.

\bibitem[Lorch et~al.(2022)Lorch, Sussex, Rothfuss, Krause, and Sch{\"{o}}lkopf]{Lorch2022AmortizedLearning}
Lars Lorch, Scott Sussex, Jonas Rothfuss, Andreas Krause, and Bernhard Sch{\"{o}}lkopf.
\newblock {Amortized Inference for Causal Structure Learning}.
\newblock In \emph{Advances in Neural Information Processing Systems}, pages 13104--13118. Curran Associates, Inc., 2022.

\bibitem[Pearl(2009)]{Pearl2009Causality}
Judea Pearl.
\newblock \emph{{Causality}}.
\newblock Cambridge university press, 2009.

\bibitem[Peters et~al.(2017)Peters, Janzing, and Sch{\"{o}}lkopf]{Peters2017ElementsAlgorithms}
Jonas Peters, Dominik Janzing, and Bernhard Sch{\"{o}}lkopf.
\newblock \emph{{Elements of causal inference: foundations and learning algorithms}}.
\newblock The MIT Press, 2017.

\bibitem[Plaat(2022)]{Plaat2022DeepLearning}
Aske Plaat.
\newblock \emph{{Deep reinforcement learning}}.
\newblock Springer, 2022.

\bibitem[Ramsey and Andrews(2023)]{Ramsey2023Py-TetradSearch}
Joseph Ramsey and Bryan Andrews.
\newblock {Py-Tetrad and RPy-Tetrad: A New Python Interface with R Support for Tetrad Causal Search}.
\newblock \emph{In Causal Analysis Workshop Series PMLR}, pages 40--51, 2023.

\bibitem[Sauter et~al.(2023)Sauter, Acar, and Francois-Lavet]{Sauter2023ADiscovery}
Andreas W~M Sauter, Erman Acar, and Vincent Francois-Lavet.
\newblock {A Meta-Reinforcement Learning Algorithm for Causal Discovery}.
\newblock In \emph{Proceedings of the Second Conference on Causal Learning and Reasoning}, pages 602--619. PMLR, 2023.

\bibitem[Sauter et~al.(2024)Sauter, Botteghi, Acar, and Plaat]{Sauter2024CORE:Learning}
Andreas W~M Sauter, Nicolò Botteghi, Erman Acar, and Aske Plaat.
\newblock {CORE: Towards Scalable and Efficient Causal Discovery with Reinforcement Learning}.
\newblock \emph{Proceedings of the 23rd International Conference on Autonomous Agents and Multiagent Systems (AAMAS)}, pages 1664--1672, 2024.

\bibitem[Scheines et~al.(1998)Scheines, Spirtes, Glymour, Meek, and Richardson]{Scheines1998TheSpecification}
Richard Scheines, Peter Spirtes, Clark Glymour, Christopher Meek, and Thomas Richardson.
\newblock {The TETRAD project: Constraint based aids to causal model specification}.
\newblock \emph{Multivariate Behavioral Research}, 33\penalty0 (1):\penalty0 65--117, 1998.

\bibitem[Sharma and Kiciman(2020)]{Sharma2020DoWhy:Inference}
Amit Sharma and Emre Kiciman.
\newblock {DoWhy: An End-to-End Library for Causal Inference}.
\newblock 2020.

\bibitem[Spirtes et~al.(2000)Spirtes, Glymour, and Scheines]{Spirtes2000CausationSearch}
Peter Spirtes, Clark~N Glymour, and Richard Scheines.
\newblock \emph{{Causation, prediction, and search}}.
\newblock MIT press, 2000.

\bibitem[Taskesen(2020)]{Taskesen2020LearningPackage.}
Erdogan Taskesen.
\newblock {Learning Bayesian Networks with the bnlearn Python Package.}, 2020.

\bibitem[Todorov et~al.(2012)Todorov, Erez, and Tassa]{Todorov2012MuJoCo:Control}
Emanuel Todorov, Tom Erez, and Yuval Tassa.
\newblock {MuJoCo: A physics engine for model-based control}.
\newblock \emph{IEEE/RSJ international conference on intelligent robots and systems}, pages 5026--5033, 2012.

\bibitem[Toth et~al.(2022)Toth, Lorch, Knoll, Krause, Pernkopf, Peharz, and von K{\"{u}}gelgen]{Toth2022ActiveInference}
Christian Toth, Lars Lorch, Christian Knoll, Andreas Krause, Franz Pernkopf, Robert Peharz, and Julius von K{\"{u}}gelgen.
\newblock {Active Bayesian Causal Inference}.
\newblock In \emph{Advances in Neural Information Processing Systems}, pages 16261--16275. Curran Associates, Inc., 2022.

\bibitem[Towers et~al.(2023)Towers, Terry, Kwiatkowski, Balis, Cola, Deleu, Goul{\~{a}}o, Kallinteris, KG, Krimmel, Perez-Vicente, Pierr{\'{e}}, Schulhoff, Tai, Shen, and Younis]{Towers2023Gymnasium}
Mark Towers, Jordan~K Terry, Ariel Kwiatkowski, John~U Balis, Gianluca~de Cola, Tristan Deleu, Manuel Goul{\~{a}}o, Andreas Kallinteris, Arjun KG, Markus Krimmel, Rodrigo Perez-Vicente, Andrea Pierr{\'{e}}, Sander Schulhoff, Jun~Jet Tai, Andrew Tan~Jin Shen, and Omar~G Younis.
\newblock {Gymnasium}, 2023.

\bibitem[Van~den Bulcke et~al.(2006)Van~den Bulcke, Van~Leemput, Naudts, van Remortel, Ma, Verschoren, De~Moor, and Marchal]{VandenBulcke2006SynTReN:Algorithms}
Tim Van~den Bulcke, Koenraad Van~Leemput, Bart Naudts, Piet van Remortel, Hongwu Ma, Alain Verschoren, Bart De~Moor, and Kathleen Marchal.
\newblock {SynTReN: a generator of synthetic gene expression data for design and analysis of structure learning algorithms}.
\newblock \emph{BMC Bioinformatics}, 7\penalty0 (1):\penalty0 43, 2006.

\bibitem[Vowels et~al.(2023)Vowels, Camgoz, and Bowden]{Vowels23yall}
Matthew~J. Vowels, Necati~Cihan Camgoz, and Richard Bowden.
\newblock {D’ya Like DAGs? A Survey on Structure Learning and Causal Discovery}.
\newblock \emph{ACM Computing Surveys}, 55\penalty0 (4):\penalty0 1--36, 2023.

\bibitem[Zhang et~al.(2021)Zhang, Zhu, Kalander, Ng, Ye, Chen, and Pan]{Zhang2021GCastle:Discovery}
Keli Zhang, Shengyu Zhu, Marcus Kalander, Ignavier Ng, Junjian Ye, Zhitang Chen, and Lujia Pan.
\newblock {gCastle: A Python Toolbox for Causal Discovery}.
\newblock 2021.

\bibitem[Zheng et~al.(2018)Zheng, Aragam, Ravikumar, and Xing]{Zheng2018DAGsLearning}
Xun Zheng, Bryon Aragam, Pradeep Ravikumar, and Eric~P Xing.
\newblock {DAGs with NO TEARS: Continuous Optimization for Structure Learning}.
\newblock 2018.

\bibitem[Zhu et~al.(2019)Zhu, Ng, and Chen]{Zhu2019CausalLearning}
Shengyu Zhu, Ignavier Ng, and Zhitang Chen.
\newblock {Causal Discovery with Reinforcement Learning}.
\newblock \emph{ArXiv}, abs/1906.04477, 2019.

\end{thebibliography}
}

\end{document}